\documentclass[sigconf, screen, nonacm]{acmart}

\AtBeginDocument{}

\usepackage{booktabs,multirow,makecell,tabularx,array}
\usepackage{algorithm}
\usepackage{algpseudocode}
\usepackage[table]{xcolor}
\usepackage{xspace}
\definecolor{tabred}{HTML}{C83030}
\definecolor{tabblue}{HTML}{3068B0}

\graphicspath{{figures/}}

\newcommand{\Method}{CoRegOVCD\xspace}
\newcommand{\SPD}{SPD\xspace}
\newcommand{\IoUC}{IoU$_C$}
\newcommand{\bestred}[1]{\textcolor{tabred}{\textbf{#1}}}
\newcommand{\secondblue}[1]{\textcolor{tabblue}{\textbf{#1}}}
\newcommand{\FoneC}{F1$_C$}
\newcommand{\best}[1]{\textbf{#1}}

\newcommand{\metricblue}[1]{\cellcolor{blue!8}\textbf{#1}}
\newcommand{\metricorange}[1]{\cellcolor{orange!10}\textbf{#1}}
\DeclareRobustCommand{\metricorangecaption}[1]{\begingroup\setlength{\fboxsep}{1pt}\colorbox{orange!10}{\strut\textbf{#1}}\endgroup}
\newcommand{\tablesection}[2]{\multicolumn{#1}{l}{\scriptsize\textit{#2}}\\}
\newcommand{\oursrow}{\rowcolor{gray!10}}

\newcolumntype{Y}{>{\raggedright\arraybackslash}X}
\newcolumntype{C}[1]{>{\centering\arraybackslash}p{#1}}

\title{\Method: Consistency-Regularized Open-Vocabulary Change Detection}

\author{Weidong Tang}
\email{wdtang29@gmail.com}
\affiliation{
  \institution{China Agricultural University}
  \department{College of Information and Electrical Engineering}
  \city{Beijing}
  \country{China}}

\author{Hanbin Sun}
\email{2023308130201@cau.edu.cn}
\affiliation{
  \institution{China Agricultural University}
  \department{College of Information and Electrical Engineering}
  \city{Beijing}
  \country{China}}

\author{Zihan Li}
\email{2023308160205@cau.edu.cn}
\affiliation{
  \institution{China Agricultural University}
  \department{College of Information and Electrical Engineering}
  \city{Beijing}
  \country{China}}

\author{Yikai Wang}
\email{yikaiwang745@gmail.com}
\affiliation{
  \institution{China Agricultural University}
  \department{College of Information and Electrical Engineering}
  \city{Beijing}
  \country{China}}

\author{Feifan Zhang}
\authornote{Corresponding author.}
\email{feifanzhang@cau.edu.cn}
\affiliation{
  \institution{China Agricultural University}
  \department{College of Science}
  \city{Beijing}
  \country{China}}

\renewcommand{\shortauthors}{Tang et al.}

\begin{document}

\begin{abstract}
Remote sensing change detection (CD) aims to identify where land-cover semantics change across time, but most existing methods still assume a fixed label space and therefore cannot answer arbitrary user-defined queries. Open-vocabulary change detection (OVCD) instead asks for the change mask of a queried concept. In the fully training-free setting, however, dense concept responses are difficult to compare directly across dates: appearance variation, weak cross-concept competition, and the spatial continuity of many land-cover categories often produce noisy, fragmented, and semantically unreliable change evidence. We propose Consistency-Regularized Open-Vocabulary Change Detection (CoRegOVCD), a training-free dense inference framework that reformulates concept-specific change as calibrated posterior discrepancy. Competitive Posterior Calibration (CPC) and the Semantic Posterior Delta (SPD) convert raw concept responses into competition-aware queried-concept posteriors and quantify their cross-temporal discrepancy, making semantic change evidence more comparable without explicit instance matching. Geometry-Token Consistency Gate (GeoGate) and Regional Consensus Discrepancy (RCD) further suppress unsupported responses and improve spatial coherence through geometry-aware structural verification and regional consensus. Across four benchmarks spanning building-oriented and multi-class settings, CoRegOVCD consistently improves over the strongest previous training-free baseline by 2.24 to 4.98 \FoneC{} points and reaches a six-class average of 47.50\% \FoneC{} on SECOND.
\end{abstract}

\maketitle
\vspace{-0.5em}
\section{Introduction}
\begin{figure}
    \centering
    \includegraphics[width=1.0\linewidth]{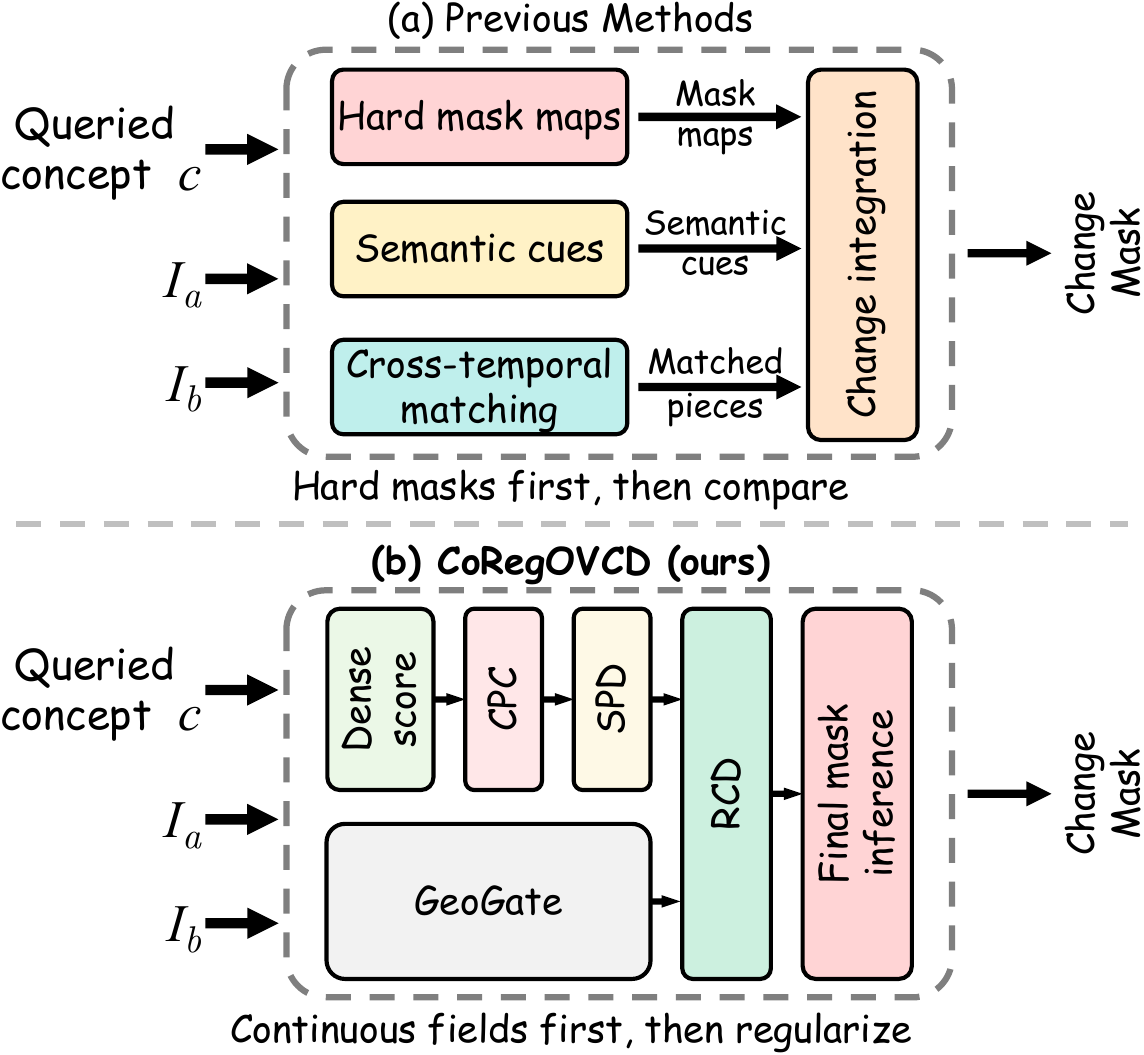}
    \vspace{-1.5em}
    \caption{Paradigm comparison between previous training-free OVCD methods and \Method. (a) Previous methods typically rely on explicit mask or instance representations, semantic cues, and cross-temporal matching before producing the final change mask. (b) \Method instead performs dense posterior-based change inference through dense score construction, CPC, SPD, GeoGate, RCD, and lightweight final mask inference.}
    \Description{A two-part flowchart comparing prior training-free OVCD paradigms with CoRegOVCD. In panel (a), a text query and two temporal images feed a multi-stage pipeline with hard mask maps, semantic cues, cross-temporal matching, and change integration, followed by a change mask. In panel (b), the same inputs feed a dense score, CPC, SPD, GeoGate, RCD, and final mask inference pipeline, which then outputs the change mask.}
    
    \label{fig:paradigm_contrast}
    \vspace{-2em}
\end{figure}
Remote sensing change detection (CD) aims to identify semantic changes from multi-temporal observations of the same geographic area \cite{yao2025mtascd,tan2025triples,tang2024clearscd,liu2025gstm,wang2024dual}, playing a crucial role in environmental monitoring, urban expansion tracking, and disaster assessment \cite{lv2025land,dou2025mcamamba,li2024comic,zheng2025semantic,singh1989review}. Current CD methods can typically be divided into two categories: binary change detection (BCD) \cite{yang2025binary}, which determines whether a change has occurred, and semantic change detection (SCD) \cite{ning2024semantic,chen2024scdvit}, which further identifies the categories of change. However, both paradigms are built upon a closed-world assumption characterized by a predefined label space, which limits their ability to generalize to unseen categories  \cite{korkmaz2026referring,zhu2025univcd}. In addition, traditional methods predominantly rely on supervised or semi-supervised paradigms, which depend on costly pixel-level annotations, making them even more impractical for open-world scenarios \cite{hou2025language,zou2026semi,peng2025toward}.

Recent advances in large-scale foundation models have opened new opportunities to address these limitations. Through extensive pretraining on massive image and language datasets, these models capture rich semantic knowledge and exhibit strong zero-shot generalization capabilities, enabling them to interpret and reason about previously unseen concepts without task-specific adaptation~\cite{zhu2024awt,bousselham2024grounding,yu2024clipceil}. Early works such as AnyChange \cite{zheng2024segment} and UCD-SCM \cite{tan2024segment} have demonstrated the potential of leveraging pre-trained foundation models for zero-shot change detection, achieving promising results under open-world conditions. Motivated by these developments, the concept of open-vocabulary change detection (OVCD) has been proposed \cite{li2025dynamicearth}, aiming to detect semantic changes across any user-specified category.

Current mainstream approaches primarily decompose the OVCD task into three subtasks, which are mask segmentation, semantic recognition, and change detection, with representative works such as DynamicEarth \cite{li2025dynamicearth} and AdaptOVCD \cite{dou2026adaptovcd}, while OmniOVCD \cite{zhang2026omniovcd} introduces a standalone architecture aimed at reducing model cascade error. However, existing methods still face two fundamental challenges: (1) most approaches rely on binary decisions for pixel-wise classification of given prompts, which can lead to misclassification when candidate concepts are semantically similar; (2) variations in external conditions (illumination, seasonal changes, and atmospheric effects) often introduce spurious differences that result in inconsistencies in cross-temporal semantics and spatial neighborhoods \cite{yang2025scednet}.
To address these challenges, our core insight is to represent target vocabulary concepts using continuous probability values rather than conventional binary masks. This design naturally captures the model's confidence in predicting target semantics, offering a more nuanced representation than binary decisions. To further suppress pseudo changes, we introduce geometric cues as a complementary constraint. Intuitively, regions exhibiting substantial contour variations are more likely to correspond to genuine changes, whereas pseudo changes often manifest with irregular or scattered boundary patterns.

Driven by the above motivation, we propose a consistency-regularized open-vocabulary change detection framework (CoRegOVCD). As illustrated in Figure~\ref{fig:paradigm_contrast}, previous training-free OVCD methods typically generate explicit mask or instance representations early and then perform cross-temporal matching, whereas \Method reasons over dense concept responses and calibrated posterior discrepancies before a lightweight final mask inference stage.
Specifically, we introduce Competitive Posterior Calibration (CPC) to model cross-concept competition and obtain calibrated posteriors, followed by Semantic Posterior Delta (SPD) to capture cross-temporal semantic discrepancies without explicit instance matching. We further introduce Geometry-Token Consistency Gate (GeoGate), a structural verification module that leverages geometry-aware representations to validate semantic evidence. Finally, Regional Consensus Discrepancy (RCD) fuses semantic and structural cues with superpixel-based regional pooling to enforce spatial coherence. The main contributions of this work are as follows:
\begin{itemize}
    \item We propose a new training-free open-vocabulary change detection framework, CoRegOVCD, which reinterprets concept-specific change detection as a dense inference framework of calibrated posterior divergence, yielding more comparable evidence of semantic changes.
    \item We establish a complementary synergy between semantic posterior divergence and geometric structural verification, enabling robust suppression of external-variation-induced false changes while preserving genuine semantic transitions.
    \item We validate the superiority of our method on four benchmark datasets, achieving new state-of-the-art performance under training-free settings while maintaining high efficiency.
\end{itemize}

\section{Related Work}
\subsection{Remote Sensing Change Detection}
Remote sensing change detection has evolved from classical discrepancy modeling to deep feature learning. Early unsupervised methods such as CVA \cite{bovolo2007cva}, IRMAD~\cite{nielsen2007irmad}, PCA-Kmeans~\cite{celik2009pcakmeans}, ISFA~\cite{wu2014isfa}, DSFA~\cite{du2019dsfa}, and DCVA~\cite{saha2019dcva} estimate change by analyzing spectral or feature-space differences without requiring pixel-level annotations. Although these methods are attractive from the standpoint of annotation cost, they primarily operate on low-level or weakly semantic change cues and therefore often respond to illumination variation, seasonal shifts, and registration noise. Recent supervised models, including Siamese CNNs \cite{daudt2018fully}, MetaChanger \cite{fang2023changer}, and HATNet \cite{xu2024hybrid}, substantially improve localization quality by learning stronger temporal representations. However, their prediction space remains fundamentally closed-set: the semantic categories that can be recognized at test time are limited by the labels available during training. This tension between semantic flexibility and annotation dependence motivates open-vocabulary formulations of change detection.

\subsection{Foundation Models for Open-Vocabulary Perception}
The emergence of foundation models has made open-vocabulary reasoning feasible in remote sensing. CLIP~\cite{radford2021learning} provides transferable vision-language alignment for text-guided recognition, while SAM~\cite{kirillov2023segment} and its successors~\cite{carion2025sam} offer promptable mask generation with increasingly strong generalization capacity. Yet directly applying foundation models trained on natural images to aerial or satellite imagery is non-trivial because of the pronounced domain gap between horizontal-view natural scenes and overhead remote sensing observations. Recent remote-sensing-specific open-vocabulary segmentation work, such as SegEarth-OV \cite{li2025segearthov}, shows that dense high-resolution features and training-free adaptation are particularly important for transferring open-vocabulary perception to remote sensing imagery. These advances provide the semantic and spatial building blocks required by OVCD, but they do not by themselves solve the temporal consistency problem: a valid OVCD model must not only recognize the queried concept, but also determine whether its spatial support changes across two acquisition times under significant appearance perturbation.

\begin{figure*}
    \centering
    \includegraphics[width=1\linewidth]{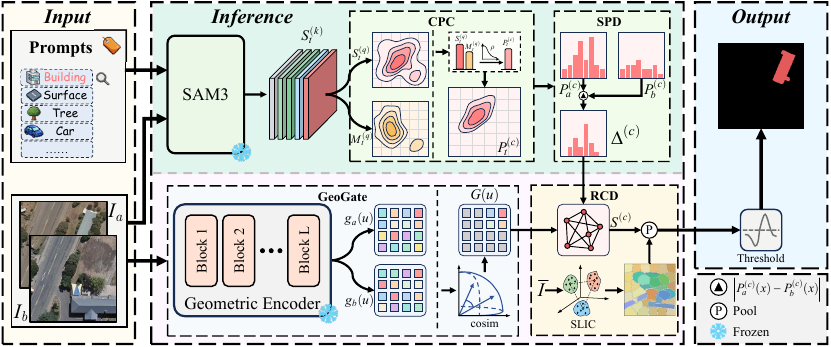}
    \caption{Overview of \Method. Given bi-temporal images and a queried concept, the framework first constructs dense concept confidence scores from prompt-conditioned Segment Anything Model~3 (SAM~3) outputs and calibrates them with Competitive Posterior Calibration (CPC). It then computes the Semantic Posterior Delta (SPD) as the semantic change signal. For structural verification, Geometry-Token Consistency Gate (GeoGate) employs a Geometric Encoder instantiated with Depth Anything~3 (DA3) to extract geometry tokens and derive the gate map $G$. Finally, Regional Consensus Discrepancy (RCD) fuses SPD and the gate map $G$ and imposes SLIC-based regional consensus, after which a lightweight final mask inference stage converts the resulting score map into the final change mask.}
    \Description{Overview diagram of CoRegOVCD. The left side shows two temporal input images and a text query. The center shows dense score construction and CPC on prompt-conditioned SAM~3 outputs, SPD computation, GeoGate built from the Geometric Encoder instantiated with Depth Anything 3 and the derived gate map, RCD with SLIC-based regional pooling, and a final mask inference stage with thresholding and structured filtering. The right side shows the final change mask.}
\vspace{-1.5em}
    \label{fig:framework}
\end{figure*}

\subsection{Training-Free Open-Vocabulary Change Detection}
Existing training-free OVCD methods can be viewed as different ways of combining foundation-model segmentation, comparison, and semantic filtering. AnyChange~\cite{zheng2024segment} uses latent matching in SAM to produce class-agnostic change masks, demonstrating the feasibility of zero-shot change localization but not category-specific reasoning. UCD-SCM~\cite{tan2024segment} adds semantic filtering on top of foundation-model proposals, but it is still largely tailored to building-oriented change detection. DynamicEarth~\cite{li2025dynamicearth} formally defines OVCD and organizes training-free inference into two modular paradigms, namely mask-compare-identify and identify-mask-compare, thereby establishing a clear benchmark and evaluation protocol for the field. AdaptOVCD~\cite{dou2026adaptovcd} strengthens this modular line through adaptive information fusion across data, feature, and decision levels, improving robustness but still relying on heterogeneous model collaboration. OmniOVCD~\cite{zhang2026omniovcd} instead exploits the decoupled outputs of SAM~3 to streamline the pipeline through instance fusion and matching in a more unified architecture. Despite these advances, existing methods still predominantly reason through explicit proposals, instance matching, or multi-stage model composition, which can be brittle to the pseudo-changes and can amplify cross-model mismatch. This gap motivates a dense, posterior-based alternative that reasons directly on concept confidence and cross-temporal geometric consistency.

\section{Methodology}
\textbf{Problem setup.}
Open-vocabulary change detection (OVCD) takes two aligned remote sensing images $I_a, I_b \in \mathbb{R}^{H \times W \times 3}$ and a queried concept $c$, and predicts a binary mask $\hat{Y}^{(c)} \in \{0,1\}^{H \times W}$ over the pixel domain $\Omega$ that marks where the membership of concept $c$ changes between the two acquisition times. In the fully training-free setting considered here, all backbones remain frozen, and the user specifies the target concept with either a single prompt or a small prompt set $\mathcal{Q}(c)$. The main difficulty is not single-date recognition alone, but constructing a query-conditioned dense signal that remains comparable across time: raw concept responses may drift under appearance variation, weak cross-class competition, and the spatial continuity of many land-cover categories, even when the underlying semantic state does not truly change.

\textbf{Pipeline overview.}
As illustrated in Figure~\ref{fig:framework}, \Method addresses this issue with four main components and a lightweight final mask inference stage. Competitive Posterior Calibration (CPC) first constructs dense concept confidence scores and calibrates them into queried-concept-vs-rest posteriors $P_a^{(c)}$ and $P_b^{(c)}$. Semantic Posterior Delta (SPD) then measures their cross-temporal discrepancy as the core semantic change signal $\Delta^{(c)}$. For structural verification, Geometry-Token Consistency Gate (GeoGate) employs a Geometric Encoder instantiated with Depth Anything~3 (DA3) to extract geometry tokens and derive the gate map $G$. Regional Consensus Discrepancy (RCD) fuses $\Delta^{(c)}$ and $G$ and regularizes the fused score through regional consensus. Finally, a lightweight final mask inference stage converts the pooled score map into the output change mask. The following subsections detail Competitive Posterior Calibration, Semantic Posterior Delta, Geometry-Token Consistency Gate, Regional Consensus Discrepancy, and the final mask inference stage.
\subsection{Semantic reliability: CPC}
For each image $I_t$ with $t \in \{a,b\}$, we construct per-pixel concept confidence scores
\begin{equation}
S_t^{(k)}(x) \in [0,1], \qquad k \in \{1,\dots,K\},
\end{equation}
over a prompt vocabulary $\mathcal{V} = \{w_k\}_{k=1}^{K}$, where $x$ denotes the pixel location. Specifically, for prompt $w_k$, let
\begin{equation}
\left\{ \big(\pi_{t,i}^{(k)}(x), \alpha_{t,i}^{(k)}\big) \right\}_{i=1}^{N_t^{(k)}}
\end{equation}
denote the retained instance masks after top-$R$ truncation and confidence filtering, where $\pi_{t,i}^{(k)}(x) \in [0,1]$ is the upsampled value of the $i$-th mask at pixel $x$, $\alpha_{t,i}^{(k)} \in [0,1]$ is its retained confidence, and $N_t^{(k)}$ is the number of retained instances. Let $d_t^{(k)}(x) \in [0,1]$ denote the upsampled dense semantic response of prompt $w_k$, and set $d_t^{(k)}(x)=0$ when this branch is unavailable. We then define
\begin{equation}
S_t^{(k)}(x) =
\max\!\left(
\max_{1 \le i \le N_t^{(k)}} \alpha_{t,i}^{(k)} \pi_{t,i}^{(k)}(x),
\; d_t^{(k)}(x)
\right),
\label{eq:score_construct}
\end{equation}
with the convention that the inner maximum is $0$ when $N_t^{(k)}=0$. Eq.~\eqref{eq:score_construct} aggregates prompt-conditioned instance evidence into a single dense response while allowing the dense semantic branch to complement regions that are not well covered by the retained instances.

Let $q$ be the index of the queried concept. Direct temporal differencing of $S_a^{(q)}$ and $S_b^{(q)}$ is unreliable because the raw score does not encode cross-concept dominance, its magnitude may drift under appearance variation, and weak off-target activations may survive into the difference map.

CPC makes the dense evidence comparable across time by modeling cross-concept competition explicitly. For the queried concept, we define the strongest non-target competitor as
\begin{equation}
M_t^{(q)}(x) = \max_{k \neq q} S_t^{(k)}(x).
\end{equation}
The calibrated posterior is then
\begin{equation}
P_t^{(c)}(x) =
S_t^{(q)}(x)
\left(
\frac{S_t^{(q)}(x)}
{S_t^{(q)}(x) + M_t^{(q)}(x) + \epsilon}
\right)^{\rho},
\label{eq:calibration}
\end{equation}
where $\epsilon > 0$ is a numerical stabilizer and $\rho$ controls how strongly the strongest competitor suppresses the queried concept.

Eq.~\eqref{eq:calibration} keeps two pieces of information at once. The first factor preserves the absolute evidence of the queried concept. The second factor measures whether this evidence remains dominant under competition. When the queried concept clearly dominates its strongest rival, the factor stays close to $1$. When the competition is tight, the calibrated response is pushed down. CPC produces competition-aware queried-concept posteriors that provide a more reliable basis for subsequent cross-temporal discrepancy estimation.

\subsection{Concept-specific change: SPD}
Once the single-date evidence has been calibrated, the next step is to compare it across time without explicit mask matching. We define the core semantic change variable as the Semantic Posterior Delta (SPD):
\begin{equation}
\Delta^{(c)}(x) = \left| P_a^{(c)}(x) - P_b^{(c)}(x) \right|.
\label{eq:spd}
\end{equation}
Under the queried-concept-vs-rest interpretation, Eq.~\eqref{eq:spd} equals the total variation distance between two Bernoulli posteriors. Accordingly, Eq.~\eqref{eq:spd} captures cross-temporal changes in concept belief rather than raw score inconsistency, which makes it a cleaner change variable for training-free OVCD. As a dense posterior discrepancy, \SPD{} does not require explicit instance correspondence.

In practice, a category may be specified by multiple prompts rather than a single query. We therefore define a prompt set
\begin{equation}
\mathcal{Q}(c) = \{p_1, p_2, \dots, p_m\}.
\end{equation}
For each prompt $p \in \mathcal{Q}(c)$, let $P_t^{(p)}(x)$ denote the calibrated posterior obtained by applying Eq.~\eqref{eq:calibration} with $p$ as the queried concept. The prompt-set aggregation is then defined by
\begin{equation}
\Delta^{(c)}(x) =
\max_{p \in \mathcal{Q}(c)}
\left| P_a^{(p)}(x) - P_b^{(p)}(x) \right|.
\label{eq:prototype}
\end{equation}
Eq.~\eqref{eq:prototype} pools prompt-level deltas over the specified prompt set and keeps the strongest prompt-aligned response when semantically related prompts are not equally reliable across scenes.

\subsection{Structural verification: GeoGate}
Even after CPC, \SPD{} may still respond to non-semantic appearance variation. Illumination shifts, shadows, seasonal texture changes, and radiometric offsets can perturb the semantic posterior without corresponding to a real land-cover transition. We therefore introduce GeoGate as a structural-verification module that validates semantic change evidence with geometry-aware cues.

GeoGate instantiates the Geometric Encoder with Depth Anything~3 (DA3)~\cite{lin2025depthanything3}. Rather than using the final scalar depth prediction, GeoGate extracts the last-layer encoder tokens from the two dates as geometry-aware representations, since these tokens retain richer local structure and contextual information for structural verification. Let
\begin{equation}
\mathbf{g}_a(u), \mathbf{g}_b(u) \in \mathbb{R}^{D}
\end{equation}
denote the geometry tokens at token-grid location $u$ for the two dates. GeoGate then constructs a geometry-token consistency gate via the cosine distance
\begin{equation}
G(u) = \frac{1}{2}\left(1 - \operatorname{cosim}(\mathbf{g}_a(u), \mathbf{g}_b(u))\right),
\label{eq:gate}
\end{equation}
where
\begin{equation}
\operatorname{cosim}(\mathbf{g}_a, \mathbf{g}_b) =
\frac{\mathbf{g}_a^\top \mathbf{g}_b}
{\|\mathbf{g}_a\|_2 \, \|\mathbf{g}_b\|_2}.
\end{equation}
After bilinear upsampling, the resulting gate map $G(x) \in [0,1]$ is small in structurally stable regions and large where the geometry representations disagree across time.

In this way, GeoGate provides structure-based verification for the semantic signal: it reinforces change evidence supported by structural variation, suppresses responses unsupported by geometry, and compensates regions where semantic evidence is weak but structural change is pronounced.

\subsection{Regional consensus: RCD}
After semantic calibration and structural verification, we still need to turn these cues into a spatially coherent score map. A direct pixel-wise fusion is often too fragmented for region-like categories and too sensitive to local noise near object boundaries. Regional Consensus Discrepancy (RCD) therefore regularizes the fused evidence with shared regional support.

RCD integrates SPD and the GeoGate output via a gated fusion formulation:
\begin{equation}
\begin{aligned}
S^{(c)}(x) ={}& \Delta^{(c)}(x)\big((1-\beta) + \beta G(x)^{\gamma}\big) \\
&+ \alpha G(x)^{\gamma},
\end{aligned}
\label{eq:fusion}
\end{equation}
where $\beta \in [0,1]$ controls the strength of gated verification, $\gamma \geq 0$ modulates the response profile of the GeoGate output, and $\alpha \geq 0$ governs the additive compensation term. The gated term preserves semantic discrepancy as the primary cue while suppressing regions not supported by geometric evidence. The additive term compensates regions with salient structural variation but weak semantic response. Using only the gated term would be overly conservative when semantic evidence is weak, whereas a purely additive design would dilute concept specificity.

The fused score in Eq.~\eqref{eq:fusion} is still pixel-wise and may exceed $1$ because of the additive compensation term. We therefore clip it to the unit interval before regional pooling:
\begin{equation}
\bar{S}^{(c)}(x) = \min\!\left(1, S^{(c)}(x)\right).
\label{eq:clip}
\end{equation}
Land-cover changes typically exhibit local spatial coherence rather than isolated pixel-wise variation. We therefore impose regional consensus using Simple Linear Iterative Clustering (SLIC) superpixels~\cite{achanta2012slic} computed on the average image
\begin{equation}
\bar{I} = \frac{1}{2}(I_a + I_b).
\end{equation}
Using the average image yields a shared local partition that is less biased toward either date alone. Let $L(x)$ denote the SLIC label assigned to pixel $x$, and let $\Omega_{L(x)}$ denote the corresponding superpixel. We pool the clipped score by
\begin{equation}
\tilde{S}^{(c)}(x) =
\frac{1}{|\Omega_{L(x)}|}
\sum_{y \in \Omega_{L(x)}} \bar{S}^{(c)}(y).
\label{eq:pooling}
\end{equation}
This step does not introduce new evidence. Instead, it enforces local agreement within a shared partition before final mask inference.

\subsection{Final mask inference}
RCD yields a region-consistent score map, but downstream evaluation requires a binary change mask. We therefore use a lightweight decoding step that converts the pooled score into the final prediction while preserving the spatial support established above.

We quantize the pooled score to 8-bit space and apply a fixed threshold:
\begin{equation}
\tilde{S}^{(c)}_{u8}(x) = \lfloor 255 \tilde{S}^{(c)}(x) \rfloor,
\end{equation}
\begin{equation}
Y_0^{(c)}(x) =
\mathbb{I}\!\left[\tilde{S}^{(c)}_{u8}(x) > \tau_{u8}\right].
\label{eq:threshold}
\end{equation}
We then apply a lightweight structural filter
\begin{equation}
\hat{Y}^{(c)} = \operatorname{StructFilter}\!\left(Y_0^{(c)}\right),
\label{eq:structfilter}
\end{equation}
where $\mathbb{I}[\cdot]$ denotes the indicator function, $\tau_{u8} \in [0,255]$ is a fixed threshold in the 8-bit domain, and $\operatorname{StructFilter}(\cdot)$ denotes lightweight morphology and connected-component filtering. This decoding stage does not inject new semantic evidence; it removes isolated fragments and stabilizes compact regions after CPC, SPD, GeoGate, and RCD have already determined the underlying score landscape. Algorithm~\ref{alg:coregovcd} summarizes the complete inference procedure.

\begin{table*}[t]
\centering
\caption{Quantitative comparison on three building-oriented benchmarks and SECOND-Building under the per-class OR-rule evaluation protocol. For SECOND-Building, a~pixel is positive if the queried class appears at either timestamp. ``-'' denotes values not numerically reported in the corresponding paper. The \bestred{best} and \secondblue{second-best} results are highlighted. \metricorangecaption{\FoneC{}} serves as the primary evaluation metric.
}
\label{tab:binary}
\setlength{\tabcolsep}{3.6pt}
\renewcommand{\arraystretch}{1.10}
\resizebox{\textwidth}{!}{
\begin{tabular}{l|cccc|cccc|cccc|cccc}
\toprule
& \multicolumn{4}{c|}{\textbf{LEVIR-CD (\%)}} & \multicolumn{4}{c|}{\textbf{WHU-CD-256 (\%)}} & \multicolumn{4}{c|}{\textbf{DSIFN (\%)}} & \multicolumn{4}{c}{\textbf{SECOND (Building) (\%)}} \\
\textbf{Method} & \metricblue{Prec.} & \metricblue{Rec.} & \metricblue{\IoUC{}} & \metricorange{\FoneC{}} & \metricblue{Prec.} & \metricblue{Rec.} & \metricblue{\IoUC{}} & \metricorange{\FoneC{}} & \metricblue{Prec.} & \metricblue{Rec.} & \metricblue{\IoUC{}} & \metricorange{\FoneC{}} & \metricblue{Prec.} & \metricblue{Rec.} & \metricblue{\IoUC{}} & \metricorange{\FoneC{}} \\
\midrule \noalign{\vskip -0.35em}
\tablesection{17}{Traditional unsupervised methods}
CVA~\cite{bovolo2007cva} & 5.43 & \secondblue{93.93} & 5.41 & 10.26 & 3.73 & \secondblue{92.20} & 3.71 & 7.16 & 18.02 & \secondblue{92.16} & 17.75 & 30.14 & 11.65 & \secondblue{97.08} & 11.61 & 20.81 \\
IRMAD~\cite{nielsen2007irmad} & 11.06 & 14.62 & 6.72 & 12.59 & 4.19 & 13.64 & 3.31 & 6.41 & 28.63 & 7.05 & 5.99 & 11.31 & 23.26 & 22.60 & 12.95 & 22.93 \\
PCA-Kmeans~\cite{celik2009pcakmeans} & 5.92 & 36.13 & 5.36 & 10.18 & 7.24 & 44.72 & 6.64 & 12.46 & 27.28 & 43.16 & 20.07 & 33.43 & 17.74 & 44.32 & 14.51 & 25.34 \\
ISFA~\cite{wu2014isfa} & 5.37 & \bestred{96.27} & 5.36 & 10.17 & 3.83 & \bestred{92.99} & 3.82 & 7.36 & 17.51 & \bestred{96.81} & 17.41 & 29.66 & 11.46 & \bestred{98.49} & 11.44 & 20.52 \\
DSFA~\cite{du2019dsfa} & 8.90 & 41.75 & 7.92 & 14.67 & 6.67 & 33.99 & 5.91 & 11.15 & 26.98 & 35.91 & 18.21 & 30.81 & 20.77 & 43.48 & 16.35 & 28.11 \\
DCVA~\cite{saha2019dcva} & 18.72 & 35.46 & 13.28 & 24.50 & 21.35 & 42.18 & 15.62 & 28.36 & 32.47 & 28.63 & 17.85 & 30.42 & 25.18 & 38.74 & 18.23 & 30.52 \\
\midrule \noalign{\vskip -0.35em}
\tablesection{17}{Training-free methods}
AnyChange~\cite{zheng2024segment} & 11.96 & 18.50 & 7.85 & 14.56 & 11.13 & 17.80 & 7.30 & 13.59 & 16.49 & 31.50 & 12.14 & 21.64 & 11.54 & 17.68 & 7.51 & 13.96 \\
UCD-SCM~\cite{tan2024segment} & 23.99 & 48.73 & 19.15 & 32.15 & 23.40 & 73.47 & 21.58 & 35.50 & 55.06 & 16.30 & 14.38 & 25.15 & 31.99 & 40.61 & 21.79 & 35.79 \\
DynamicEarth (IMC)~\cite{li2025dynamicearth} & \secondblue{66.26} & 73.52 & 53.50 & 69.70 & 70.29 & 74.86 & 56.80 & 72.50 & 48.23 & 37.54 & 26.13 & 42.18 & 47.81 & 37.45 & 26.50 & 42.00 \\
DynamicEarth (MCI)~\cite{li2025dynamicearth} & 48.36 & 60.11 & 36.60 & 53.60 & 43.90 & 84.15 & 40.60 & 57.70 & \secondblue{58.52} & 51.76 & 37.89 & 54.98 & 59.52 & 51.46 & 38.10 & 55.20 \\
AdaptOVCD~\cite{dou2026adaptovcd} & 62.83 & 74.10 & 51.52 & 68.00 & \secondblue{80.60} & 72.85 & 61.99 & 76.53 & 55.23 & 64.42 & \secondblue{42.32} & \secondblue{59.47} & \secondblue{64.38} & 63.25 & \secondblue{46.85} & \secondblue{63.81} \\
OmniOVCD~\cite{zhang2026omniovcd} & - & - & \secondblue{67.20} & \secondblue{80.40} & - & - & \secondblue{66.50} & \secondblue{79.90} & - & - & - & - & - & - & 45.20 & 62.30 \\
\oursrow\best{CoRegOVCD (Ours)} & \bestred{79.72} & 86.57 & \bestred{70.95} & \bestred{83.01} & \bestred{87.86} & 77.12 & \bestred{69.70} & \bestred{82.14} & \bestred{62.44} & 66.60 & \bestred{47.55} & \bestred{64.45} & \bestred{79.24} & 56.10 & \bestred{48.91} & \bestred{65.69} \\
\bottomrule
\end{tabular}
}
\vspace{-1.5em}
\end{table*}

\begin{algorithm}[t]
\caption{Inference procedure of \Method}
\label{alg:coregovcd}
\small
\begin{algorithmic}[1]
\Statex \textbf{Input:} Bi-temporal images $I_a$, $I_b \in \mathbb{R}^{H \times W \times 3}$; queried concept $c$; prompt set $\mathcal{Q}(c)$
\Statex \textbf{Output:} Change mask $\hat{Y}^{(c)} \in \{0,1\}^{H \times W}$
\State construct dense concept scores $\{S_a^{(k)}\}_{k=1}^{K}$ and $\{S_b^{(k)}\}_{k=1}^{K}$ by Eq.~\eqref{eq:score_construct}

\ForAll{$p \in \mathcal{Q}(c)$}
    \State $P_a^{(p)}, P_b^{(p)} \gets$ Eq.~\eqref{eq:calibration}, \quad $\Delta^{(p)} \gets \left|P_a^{(p)} - P_b^{(p)}\right|$
\EndFor
\State $\Delta^{(c)} \gets \max_{p \in \mathcal{Q}(c)} \Delta^{(p)}$ \Comment{Eq.~\eqref{eq:spd} or Eq.~\eqref{eq:prototype}}
\State $G \gets$ Eq.~\eqref{eq:gate}, \quad $S^{(c)} \gets$ Eq.~\eqref{eq:fusion}, \quad $\bar{S}^{(c)} \gets$ Eq.~\eqref{eq:clip}
\State $L \gets \operatorname{SLIC}\!\left((I_a + I_b)/2\right)$, \quad $\tilde{S}^{(c)} \gets$ Eq.~\eqref{eq:pooling} on $L$
\State $Y_0^{(c)} \gets \mathbb{I}\!\left[\lfloor 255 \tilde{S}^{(c)} \rfloor > \tau_{u8}\right]$ \Comment{Eq.~\eqref{eq:threshold}}
\State $\hat{Y}^{(c)} \gets \operatorname{StructFilter}\!\left(Y_0^{(c)}\right)$ \Comment{Eq.~\eqref{eq:structfilter}}

\State \Return $\hat{Y}^{(c)}$
\end{algorithmic}
\end{algorithm}

\section{Experiments and Analysis}

\subsection{Datasets}
To fully assess the proposed methods, experiments are conducted on four diverse datasets: LEVIR-CD~\cite{chen2020levir}, WHU-CD-256~\cite{ji2018whucd}, DSIFN~\cite{zhang2020dsifn}, and the SECOND validation split~\cite{yang2021second} for multi-class semantic change detection. LEVIR-CD contains 637 bi-temporal image pairs of size 1024$\times$1024 at 0.5 m resolution and mainly focuses on building changes in urban scenes. WHU-CD-256 is derived from very-high-resolution aerial imagery and cropped into 256$\times$256 patches, providing a challenging benchmark for building change detection in dense urban areas. DSIFN contains 3940 bi-temporal image pairs of size 512$\times$512 at approximately 2 m resolution, covering diverse scenes and building-related changes across different imaging conditions. SECOND contains 4662 bi-temporal image pairs of size 512$\times$512 with pixel-wise semantic annotations for six land-cover categories, making it a suitable benchmark for evaluating semantic change detection beyond the building-only setting.

\subsection{Experimental Setup}

\begin{table*}[t]
\centering
\caption{Open-vocabulary change detection results on the six SECOND classes under the per-class OR-rule evaluation, where a~pixel is positive for class $c$ if either timestamp belongs to $c$. The \bestred{best} and \secondblue{second-best} results are highlighted. \metricorangecaption{\FoneC{}} serves as the primary evaluation metric.}
\vspace{-0.5em}
\label{tab:second}
\setlength{\tabcolsep}{3.7pt}
\resizebox{\textwidth}{!}{
\begin{tabular}{l|cc|cc|cc|cc|cc|cc|cc}
\toprule
& \multicolumn{2}{c|}{\textbf{Building (\%)}} & \multicolumn{2}{c|}{\textbf{Tree (\%)}} & \multicolumn{2}{c|}{\textbf{Water (\%)}} & \multicolumn{2}{c|}{\textbf{Low vegetation (\%)}} & \multicolumn{2}{c|}{\textbf{Surface (\%)}} & \multicolumn{2}{c|}{\textbf{Playground (\%)}} & \multicolumn{2}{c}{\textbf{Class Avg. (\%)}} \\
\textbf{Method} & \metricblue{\IoUC{}} & \metricorange{\FoneC{}} & \metricblue{\IoUC{}} & \metricorange{\FoneC{}} & \metricblue{\IoUC{}} & \metricorange{\FoneC{}} & \metricblue{\IoUC{}} & \metricorange{\FoneC{}} & \metricblue{\IoUC{}} & \metricorange{\FoneC{}} & \metricblue{\IoUC{}} & \metricorange{\FoneC{}} & \metricblue{\IoUC{}} & \metricorange{\FoneC{}} \\
\midrule
DynamicEarth (IMC)~\cite{li2025dynamicearth} & 26.50 & 42.00 & 13.50 & 23.80 & 9.80 & 17.90 & 0.00 & 0.00 & 0.00 & 0.00 & 16.50 & 28.30 & 11.05 & 18.67 \\
DynamicEarth (MCI)~\cite{li2025dynamicearth} & 38.10 & 55.20 & \secondblue{20.30} & \secondblue{33.80} & 14.30 & 25.10 & 24.10 & 38.90 & 26.20 & 41.60 & 20.00 & 33.30 & 23.83 & 37.98 \\
AdaptOVCD~\cite{dou2026adaptovcd} & \secondblue{46.85} & \secondblue{63.81} & 12.67 & 22.49 & \secondblue{23.34} & \secondblue{37.84} & 22.15 & 36.27 & \bestred{33.99} & \bestred{50.74} & \secondblue{29.28} & \secondblue{45.30} & \secondblue{28.05} & \secondblue{42.74} \\
OmniOVCD~\cite{zhang2026omniovcd} & 45.20 & 62.30 & 16.70 & 28.60 & 21.20 & 35.00 & \secondblue{24.50} & \secondblue{39.30} & 27.70 & 43.40 & 27.00 & 42.40 & 27.10 & 41.80 \\
\oursrow\best{CoRegOVCD (Ours)} & \bestred{48.91} & \bestred{65.69} & \bestred{20.93} & \bestred{34.61} & \bestred{29.88} & \bestred{46.01} & \bestred{27.71} & \bestred{43.40} & \secondblue{32.27} & \secondblue{48.79} & \bestred{30.31} & \bestred{46.52} & \bestred{31.67} & \bestred{47.50} \\
\bottomrule
\end{tabular}
}

\end{table*}

\textbf{Metrics.} All metrics are computed on the changed class. We report changed-class precision (\%), recall (\%), \IoUC{} (\%), and \FoneC{} (\%). We treat \FoneC{} as the primary ranking metric because it balances precision and recall, while \IoUC{} is reported as a complementary overlap measure.

\textbf{Implementation details.} All experiments are conducted on a single NVIDIA A800-SXM4-80GB GPU under PyTorch. All backbones remain frozen. CPC is built on dense concept confidence scores constructed from prompt-conditioned SAM~3 outputs. GeoGate uses DA3 as the Geometric Encoder, and its geometry tokens are extracted from feature layer~23 at a processing resolution of 336. Unless noted otherwise, the calibration exponent is $\rho=1.5$, the fusion parameters are $\alpha=0.1$, $\beta=0.7$, and $\gamma=1$, the concept confidence threshold is 0.5, and the retained instance set is truncated to top-$R=30$. The 8-bit thresholds, SLIC segment counts, and lightweight decoding hyperparameters are fixed within each evaluation configuration. We use a fixed category-specific prompt configuration across experiments. No fine-tuning, adapter training, or dataset-specific backbone update is performed. Additional implementation details are included in the \textit{Supplement}.

\textbf{Compared methods.} We compare against traditional unsupervised baselines (CVA~\cite{bovolo2007cva}, IRMAD~\cite{nielsen2007irmad}, PCA-Kmeans~\cite{celik2009pcakmeans}, ISFA~\cite{wu2014isfa}, DSFA~\cite{du2019dsfa}, and DCVA~\cite{saha2019dcva}), early training-free methods (AnyChange~\cite{zheng2024segment} and UCD-SCM~\cite{tan2024segment}), and recent OVCD systems including DynamicEarth~\cite{li2025dynamicearth}, AdaptOVCD~\cite{dou2026adaptovcd}, and OmniOVCD~\cite{zhang2026omniovcd}. We report DynamicEarth (MCI) and DynamicEarth (IMC), which correspond to the SAM-DINOv2-SegEarth-OV and APE-/-DINO configurations, respectively.

\subsection{Comparison with State-of-the-Art Methods}
\textbf{Building-oriented benchmarks.} Table~\ref{tab:binary} compares \Method with prior methods on LEVIR-CD, WHU-CD-256, DSIFN, and the building split of SECOND. \Method achieves the best \FoneC{} on all four settings, surpassing the strongest training-free baseline by 2.61, 2.24, 4.98, and 1.88 points on LEVIR-CD, WHU-CD-256, DSIFN, and SECOND-Building, respectively. Traditional unsupervised baselines typically exhibit very high recall but extremely low precision, showing that purely discrepancy-driven methods overreact to non-semantic appearance variation. Recent training-free OVCD baselines improve semantic selectivity, but their outputs remain vulnerable to unstable semantic responses, insufficient structural verification, or weak regional consistency. The consistent gains across all four settings indicate that CPC-based posterior differencing, together with GeoGate, RCD, and the final decoding stage, provides a robust dense formulation for building-oriented OVCD.

\textbf{Class-wise results on SECOND.} Table~\ref{tab:second} reports class-wise results on the six queried classes of SECOND. Compared with AdaptOVCD, \Method raises the class-average \IoUC{} from 28.05\% to 31.67\% and \FoneC{} from 42.74\% to 47.50\%. The gains are especially pronounced for \emph{tree} and \emph{water}, two categories for which appearance can vary substantially while geometry remains informative for structural verification. \Method also performs best on \emph{building}, \emph{low vegetation}, and \emph{playground}; the \emph{playground} result is consistent with max-over-prompts aggregation over visually diverse sports-field instances. Overall, the class-wise comparison shows that the proposed dense posterior formulation remains effective across both object-like and region-like categories.

\subsection{Ablation Study and Analysis}
\begin{table}[t]
\centering
\caption{Ablation of CPC, GeoGate, RCD, and final mask inference in \Method. ``RCD (\emph{w/o} additive term)'' removes the additive compensation term in Eq.~\eqref{eq:fusion}, and ``Decoding (\emph{w/o} StructFilter)'' skips the structural filtering step in final mask inference. SECOND Avg. reports the class-average result over the six semantic categories.}
\label{tab:ablation}
\renewcommand{\arraystretch}{1.10}
\resizebox{\columnwidth}{!}{
\begin{tabular}{l|cc|cc|cc}
\toprule
& \multicolumn{2}{c|}{\textbf{DSIFN (\%)}} & \multicolumn{2}{c|}{\textbf{LEVIR-CD (\%)}} & \multicolumn{2}{c}{\textbf{SECOND Avg. (\%)}} \\
\textbf{Setting} & \metricblue{\IoUC{}} & \metricorange{\FoneC{}} & \metricblue{\IoUC{}} & \metricorange{\FoneC{}} & \metricblue{\IoUC{}} & \metricorange{\FoneC{}} \\
\midrule
\emph{w/o} CPC & 45.66 & 62.70 & 70.30 & 82.56 & 31.05 & 46.65 \\
\emph{w/o} GeoGate & 34.51 & 51.31 & 50.01 & 66.67 & 22.40 & 36.21 \\
RCD (\emph{w/o} additive term)
 & 23.26 & 37.74 & 69.92 & 82.05 & 31.16 & 46.85 \\
RCD (\emph{w/o} SLIC) & 44.85 & 61.93 & 68.89 & 81.32 & 29.60 & 45.36 \\
Decoding (\emph{w/o} StructFilter)
 & 42.32 & 59.47 & 70.15 & 82.46 & 31.12 & 46.81 \\
\oursrow\best{CoRegOVCD} & \best{47.55} & \best{64.45} & \best{70.95} & \best{83.01} & \best{31.67} & \best{47.50} \\
\bottomrule
\end{tabular}
}
\end{table}
\textbf{Components ablation.} Table~\ref{tab:ablation} ablates CPC, GeoGate, and the design choices in the downstream regularization and decoding stages. Removing CPC causes a consistent degradation across all three benchmarks, for example reducing \FoneC{} from 64.45\% to 62.70\% on DSIFN and from 47.50\% to 46.65\% on SECOND Avg., indicating that stable semantic change inference requires calibrated concept posteriors rather than raw concept responses alone. GeoGate is even more critical: without structural verification, \FoneC{} drops from 64.45\% to 51.31\% on DSIFN and from 47.50\% to 36.21\% on SECOND Avg., showing that posterior discrepancies must be validated by geometry-aware evidence. Within RCD, removing the additive compensation term especially hurts DSIFN, where \FoneC{} falls from 64.45\% to 37.74\%, indicating that additive compensation is particularly useful when semantic evidence is weak but structural variation remains salient. Removing SLIC pooling or the final structural filter also causes consistent degradation, showing that regional consensus and final mask inference further improve the spatial completeness and stability of the output mask.

\begin{table}[b]
\centering
\caption{Comparison of GeoGate designs. ``Depth'' uses only a depth-based gate from DA3, whereas ``Hybrid'' combines the depth-based gate with the geometry-token consistency gate. SECOND Avg. reports the class-average result over the six semantic categories.}
\label{tab:gate}
\renewcommand{\arraystretch}{1.10}
\resizebox{\columnwidth}{!}{
\begin{tabular}{l|cc|cc|cc}
\toprule
& \multicolumn{2}{c|}{\textbf{DSIFN (\%)}} & \multicolumn{2}{c|}{\textbf{LEVIR-CD (\%)}} & \multicolumn{2}{c}{\textbf{SECOND Avg. (\%)}} \\
\textbf{Gate} & \metricblue{\IoUC{}} & \metricorange{\FoneC{}} & \metricblue{\IoUC{}} & \metricorange{\FoneC{}} & \metricblue{\IoUC{}} & \metricorange{\FoneC{}} \\
\midrule
None & 34.51 & 51.31 & 50.01 & 66.67 & 22.40 & 36.21 \\
Depth & 19.58 & 32.75 & 46.12 & 63.13 & 22.12 & 35.29 \\
Hybrid & 25.60 & 40.76 & 66.89 & 80.16 & 26.58 & 40.72 \\
\oursrow\best{GeoGate} & \best{47.55} & \best{64.45} & \best{70.95} & \best{83.01} & \best{31.67} & \best{47.50} \\
\bottomrule
\end{tabular}
}
\end{table}

\begin{figure}[t]
    \centering
    \includegraphics[width=1\linewidth]{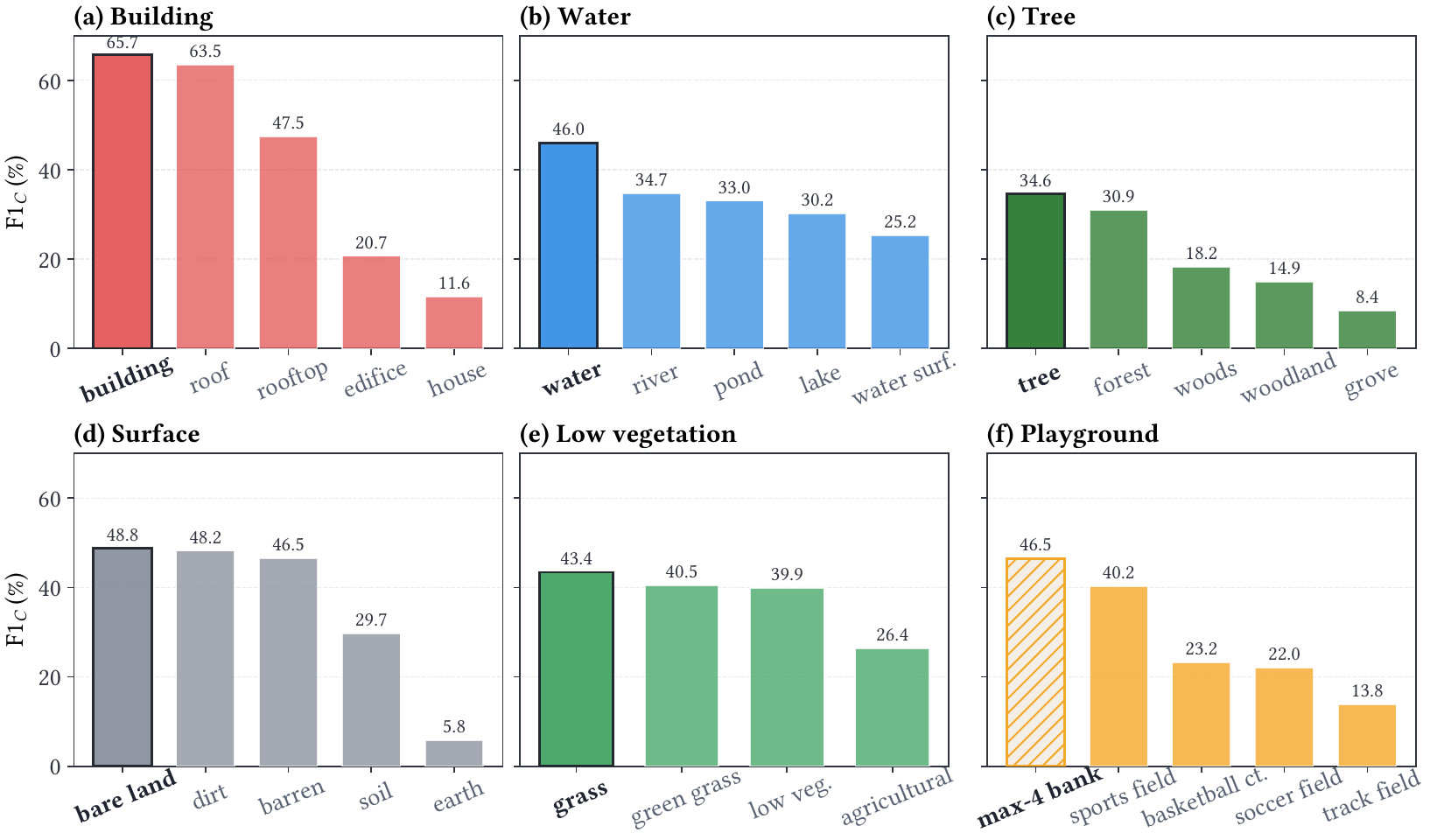}
    \caption{Open-vocabulary query substitution on SECOND. Each panel reports \FoneC{} (\%) for semantically related query words of one class. Bold bars denote the default query, and the hatched bar denotes a prompt set.}
    \Description{fig_S3_prompt_substitution}
    \label{fig:prompt_substitution}
\end{figure}

\textbf{GeoGate design.} Table~\ref{tab:gate} isolates GeoGate under fixed \SPD{} inference and RCD settings. GeoGate yields the strongest results not only on DSIFN and LEVIR-CD but also on the SECOND class average. In particular, replacing GeoGate with no structural verification causes the SECOND average to drop from 31.67\% to 22.40\% in \IoUC{} and from 47.50\% to 36.21\% in \FoneC{}, while a depth-only variant performs even worse than the no-GeoGate baseline on DSIFN and LEVIR-CD. The hybrid variant partially recovers performance, but it still trails GeoGate by a clear margin. These results show that the improvement comes from the geometry-token consistency gate rather than from simply attaching an auxiliary geometric cue.

\begin{figure}[b]
\centering
\includegraphics[width=1\columnwidth]{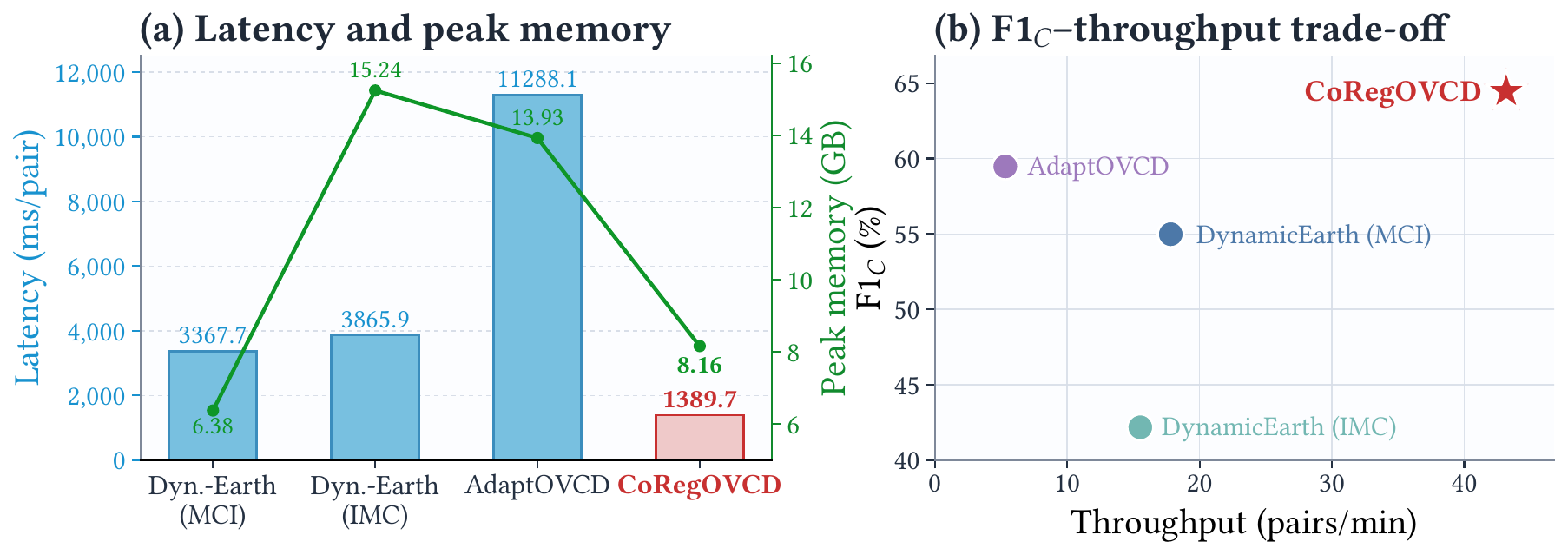}
\caption{Efficiency and accuracy trade-off among training-free OVCD methods. (a) Latency per image pair and peak memory on an NVIDIA A800-SXM4-80GB device, measured after a one-pair warm-up. (b) Accuracy--speed trade-off in terms of throughput and the corresponding DSIFN \FoneC{} values used in the efficiency comparison.}
\Description{A two-panel figure comparing efficiency and accuracy of DynamicEarth MCI, DynamicEarth IMC, AdaptOVCD, and CoRegOVCD. The left panel shows bars for latency and a line for peak memory. The right panel shows a scatter plot of throughput versus DSIFN F1_C, where CoRegOVCD lies in the top-right region with both the highest throughput and highest accuracy.}
\label{fig:efficiency_tradeoff}
\end{figure}
\textbf{Open-vocabulary query substitution.} Figure~\ref{fig:prompt_substitution} further examines whether \Method can respond to semantically related lexical variants rather than a single fixed query. Overall, several alternative query words remain effective across the six SECOND classes, which supports the open-vocabulary nature of the proposed framework. At the same time, the transfer is clearly class-dependent rather than uniform. For \emph{building}, the alternatives \emph{roof} and \emph{rooftop} still recover a substantial portion of the default performance, whereas more weakly aligned terms such as \emph{house} degrade much more noticeably. A similar pattern appears for \emph{surface} and \emph{low vegetation}, where some related expressions remain competitive but others become much less reliable. These results indicate that \Method is not tied to a single hand-crafted wording, yet the semantic alignment between the query and the deployed concept space still matters. The \emph{playground} panel is especially informative: the default max-over-4 prompt bank outperforms every single sports-field prompt, which justifies the prompt-bank design used in the main results for visually diverse categories.

\textbf{Efficiency trade-off.} Because \Method avoids proposal-level matching and multi-stage instance reasoning, it retains a simpler dense inference path. Figure~\ref{fig:efficiency_tradeoff} summarizes the resulting runtime behavior by jointly visualizing latency, peak memory, throughput, and accuracy. In Figure~\ref{fig:efficiency_tradeoff}(a), \Method achieves the lowest latency among the compared training-free OVCD methods while maintaining moderate peak memory usage. In Figure~\ref{fig:efficiency_tradeoff}(b), the DSIFN \FoneC{} (\%) values used for the efficiency comparison are plotted against throughput, making the accuracy--speed trade-off explicit. \Method processes 43.17 pairs per minute, which is 2.42$\times$ faster than DynamicEarth (MCI), 2.78$\times$ faster than DynamicEarth (IMC), and 8.11$\times$ faster than AdaptOVCD. It also attains the highest DSIFN \FoneC{} (\%) among the compared training-free OVCD methods, placing it in the most favorable region of the accuracy--efficiency space.

\begin{figure}[t]
\centering
\includegraphics[width=1\linewidth] {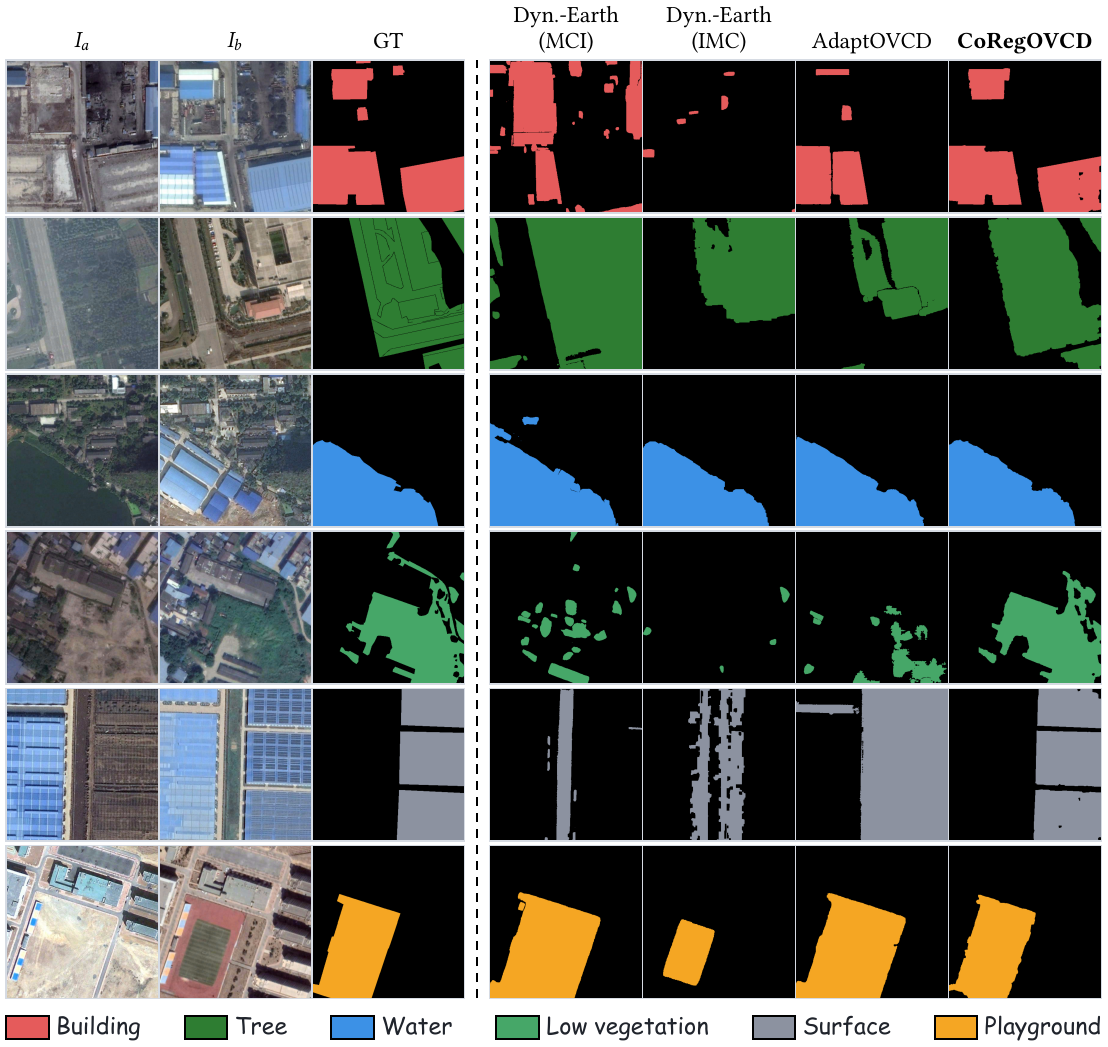}
\caption{Qualitative comparison on SECOND with DynamicEarth (MCI/IMC), AdaptOVCD, and \Method across six semantic categories. \Method yields more complete masks with fewer unsupported responses.}
\Description{Single-column qualitative comparison figure on SECOND. The columns show temporal images, ground truth, predictions from DynamicEarth MCI, DynamicEarth IMC, AdaptOVCD, and CoRegOVCD across six semantic categories.}
\label{fig:secondqual}
\end{figure}

\subsection{Qualitative Analysis}
\textbf{Qualitative comparison with prior methods.} Figure~\ref{fig:secondqual} presents qualitative comparison on six representative SECOND classes. Across building, tree, water, low vegetation, surface, and playground, \Method produces more complete masks and fewer unsupported fragments than DynamicEarth and AdaptOVCD. These visual gains are consistent with the quantitative improvements in Table~\ref{tab:second} and reflect stronger semantic selectivity, more reliable structural verification, and cleaner region-level masks.

\begin{figure}[tbp]
\centering
\includegraphics[width=\columnwidth]{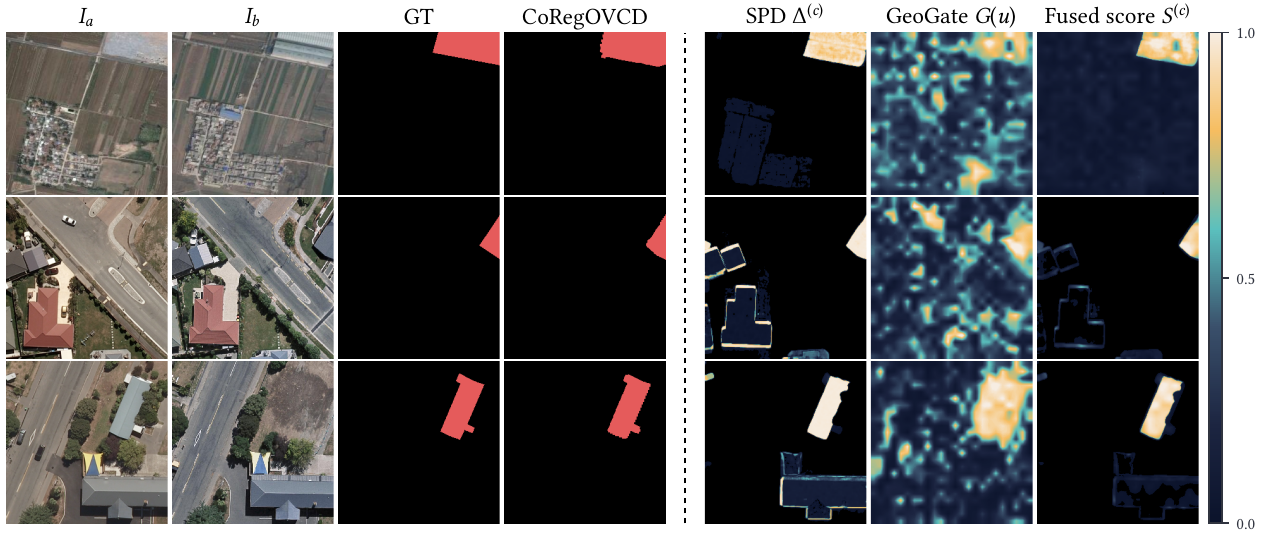}
\caption{Intermediate responses and final predictions of \Method on three representative samples. GeoGate suppresses unsupported responses, and RCD produces cleaner, more localized change evidence before final decoding.}
\Description{Single-column qualitative figure showing three representative examples with temporal images, ground truth, final prediction, the SPD map, the GeoGate output, and the RCD score map. The intermediate maps show how GeoGate suppresses noisy appearance changes before the final mask inference stage converts the RCD output into the final mask.}
\label{fig:intermediate}
\end{figure}

\textbf{Intermediate responses.} Figure~\ref{fig:intermediate} visualizes intermediate responses on three representative samples from DSIFN, LEVIR-CD, and WHU-CD-256. The \SPD{} maps already highlight the queried semantic changes, but they still contain weak responses caused by background appearance variation. The GeoGate output is spatially coarse but structurally stable, emphasizing regions that are consistent with genuine scene changes. After RCD, the score map becomes both cleaner and more localized, and the final decoding stage removes isolated fragments and stabilizes compact shapes in the output mask.

\section{Discussion}
\textbf{Why posterior differencing works.} Existing training-free OVCD pipelines usually infer change after early discretization, for example by constructing proposals, binary masks, or matched instances and then comparing them across time. Our framework follows a different inference paradigm: it retains dense concept responses, calibrates them into competition-aware queried-concept posteriors, and performs change inference directly at the posterior layer before the final binarization stage. This shift matters because it avoids committing too early to object instances or hard masks, which is especially beneficial for stuff-like categories and irregular semantic changes that do not admit reliable instance correspondence. CPC restores semantic selectivity under cross-concept competition, while GeoGate and the subsequent regional and decoding stages suppress posterior discrepancies that are not supported by structural evidence. Together, these components make posterior differencing a simpler and more stable alternative to proposal- and matching-centered training-free OVCD pipelines.

\section{Conclusion}
\Method recasts training-free open-vocabulary change detection as posterior-level dense reasoning, replacing early discretization and explicit instance matching with calibrated semantic comparison and geometry-aware structural verification. By coupling Competitive Posterior Calibration with Semantic Posterior Delta, and then regularizing the resulting change evidence through GeoGate and Regional Consensus Discrepancy, the framework makes queried-concept responses more comparable across time and yields masks that are cleaner, more complete, and less vulnerable to appearance-induced pseudo changes. The empirical results and ablation analysis show that the improvement comes less from adding a heavier pipeline than from moving change inference into a competition-aware posterior space and enforcing structural support before final binarization. This view offers a different answer to training-free OVCD: reliable open-world change analysis need not depend on increasingly elaborate proposal matching, but rather on whether semantic comparability and structural consistency are established jointly. Prompt robustness, threshold transfer, and efficient multi-concept inference remain open, but the central conclusion is clear: posterior differencing, once properly regularized, provides a stronger foundation for training-free OVCD.

\bibliographystyle{ACM-Reference-Format}
\bibliography{refs}

\clearpage

\end{document}